\documentclass[runningheads]{llncs}

\usepackage[T1]{fontenc}
\usepackage{comment}
\usepackage{graphicx}
\usepackage{enumitem}
\usepackage[hidelinks]{hyperref}
\usepackage{color}
\usepackage{bbm}
\usepackage{multirow}
\usepackage{multicol}
\usepackage{array}
\usepackage{booktabs}
\usepackage{orcidlink}
\usepackage{fontawesome5}
\usepackage[misc,geometry]{ifsym}
\usepackage{float}

\hypersetup{
  colorlinks=true,
  citecolor=blue,
  linkcolor=blue,
  urlcolor=blue
}

\begin{document}

\title{RAUNE-Net: A Residual and Attention-Driven Underwater Image Enhancement Method}
\titlerunning{RAUNE-Net: A Residual and Attention-Driven UIE Method}

\author{Wangzhen Peng\inst{1}\orcidlink{0000-0002-2333-5415}\and Chenghao Zhou\inst{2} \and Runze Hu\inst{3} \and Jingchao Cao\inst{1} \and Yutao Liu\inst{1}\textsuperscript{(\Letter)}}
\authorrunning{W. Peng et al.}
%
\institute{Department of Computer Science and Technology, Ocean University of China, Qingdao, China\\
\email{wangzhenpeng@stu.ouc.edu.cn,\{liuyutao,caojingchao\}@ouc.edu.cn} \and
School of Management Engineering, Qingdao University of Technology, China\\
\email{717946813@qq.com} \and
School of Information and Electronics, Beijing Institute of Technology, Beijing 100080, China\\
\email{hrzlpk2015@gmail.com}}

\maketitle 

\begin{abstract}
Underwater image enhancement (UIE) poses challenges due to distinctive properties of the underwater environment, including low contrast, high turbidity, visual blurriness, and color distortion. In recent years, the application of deep learning has quietly revolutionized various areas of scientific research, including UIE. However, existing deep learning-based UIE methods generally suffer from issues of weak robustness and limited adaptability. In this paper, inspired by residual and attention mechanisms, we propose a more reliable and reasonable UIE network called \textit{RAUNE-Net} by employing residual learning of high-level features at the network's bottle-neck and two aspects of attention manipulations in the down-sampling procedure. Furthermore, we collect and create two datasets specifically designed for evaluating UIE methods, which contains different types of underwater distortions and degradations. The experimental validation demonstrates that our method obtains promising objective performance and consistent visual results across various real-world underwater images compared to other eight UIE methods. Our example code and datasets are publicly available at \href{https://github.com/fansuregrin/RAUNE-Net}{https://github.com/fansuregrin/RAUNE-Net}.

\end{abstract}

\section{Introduction}
\begin{figure}
\centering
\includegraphics[width=\textwidth]{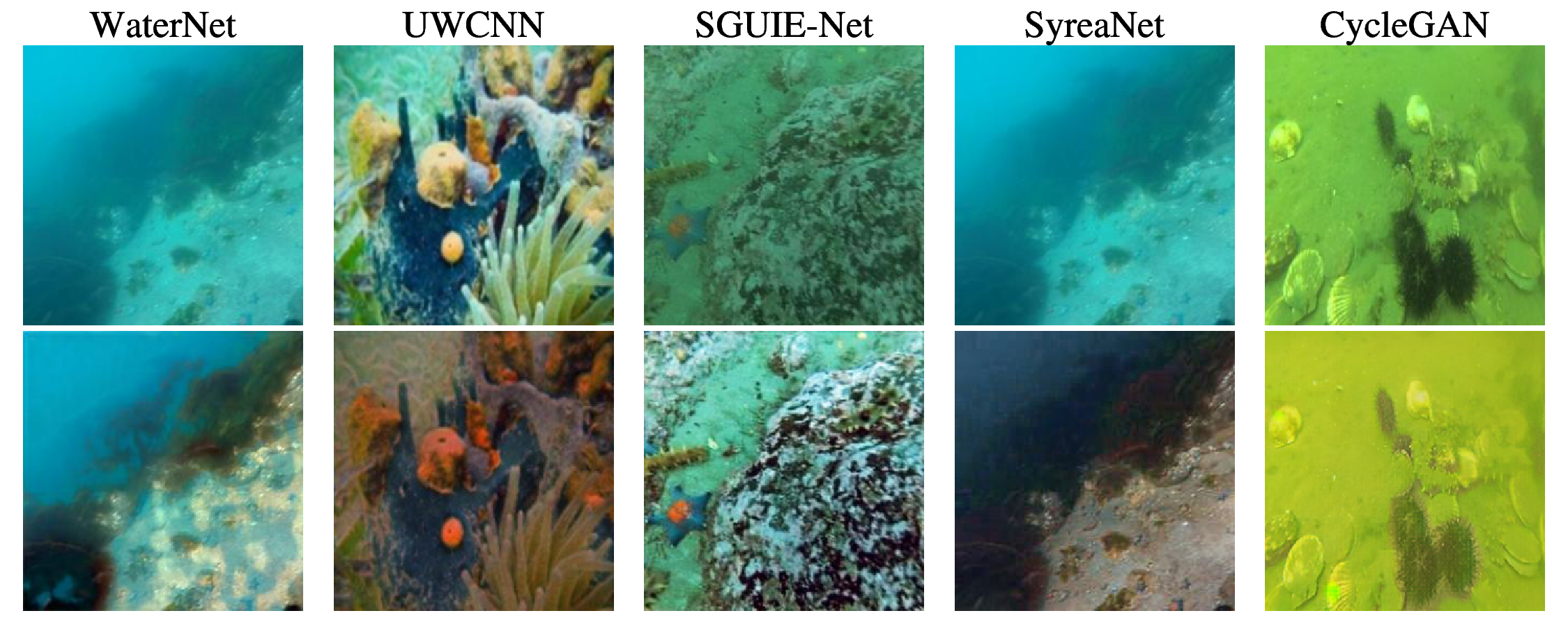}
\caption{Poor results of five UIE methods with weak robustness}
\label{5-UIE-methods}
\end{figure}

Underwater image enhancement (UIE), a challenging yet promising research direction, has garnered growing interest among researchers in recent years. Nowadays, a wide array of methods have been put forward for enhancing underwater images~\cite{liu2023uiqi}. These UIE techniques can be broadly categorized into four types~\cite{li2019underwater}: supplementary information-based, non-physical model-based, physical model-based, and data-driven methods. With the rise of deep learning and the utilization of large-scale datasets, data-driven methods have revolutionized UIE. However, it is worth noting that data-driven UIE methods still have several limitations. We have evaluated these previously proposed deep learning-based methods on multiple real-world underwater image datasets. As illustrated in Figure \hyperref[5-UIE-methods]{1}, these five methods tend to produce visually poor results when dealing with images that they are not proficient at enhancing. Therefore, further research and development are necessary to overcome these challenges and create more reliable and adaptable UIE techniques.

Based on above analysis, we propose a \textbf{R}esidual and \textbf{A}ttention-driven \textbf{UN}derwater image \textbf{E}nhancement \textbf{Net}work called \textit{RAUNE-Net} to process a wide variety of underwater images. Our main contributions can be summarized as follows:

\begin{itemize}[label=$\cdot$]
    \item We incorporate residual and attention mechanisms into UIE to construct an effective and efficient network called RAUNE-Net, which is capable of handling a wide range of underwater scenes comprehensively.
    \item Our method achieves remarkable objective performance and consistent subjective results on both referenced and non-referenced real-world underwater image testsets.
\end{itemize}

\section{Related Work}\label{sec.2}
\subsection{Convolutional Neural Network-based UIE methods}

In the early stages, Wang et al.~\cite{wang2017deep} first presented an end-to-end framework (i.e., UIE-Net) based on Convolutional Neural Network (CNN) for enhancing visual quality of underwater images. In~\cite{li2020underwater}, an underwater image dataset was synthesized firstly through an underwater image formulation algorithm based on underwater scene priors, which includes two types of water environments and five levels of distortions. Subsequently, they devised a lightweight CNN network (i.e., UWCNN) to enhance images from different underwater scenes. In contrast to aforementioned methods that used synthetic datasets for model training, Li et al.~\cite{li2019underwater} created a real-world underwater image dataset (i.e., UIEB). Then, they trained an enhancement network called Water-Net on this dataset. In detail, they generated three input images by applying white balance, histogram equalization, and gamma correction algorithms to the original underwater image. These inputs were then transformed into refined inputs using the Feature Transformation Units. Subsequently, Water-Net utilized a gated fusion network to learn three confidence maps, which were used to combine three refined inputs and produce an enhanced result. These methods either rely on synthesized images for model training or require preprocessing on raw underwater images, which still have much room for improvement.

\subsection{Generative Adversarial Network-based UIE methods}

Li et al.~\cite{li2017watergan} introduced an approach called WaterGAN, which adopted a GAN~\cite{goodfellow2020generative} to learn realistic representations of water column properties. Then, they used these generated underwater images along with their corresponding depth maps to train a color restoration network capable of real-time compensation for water column effects at a specific location. Unlike the previous method, where GAN was used solely for generating training samples, in~\cite{fabbri2018enhancing}, a GAN-based network (i.e., UGAN), which utilizes the Wasserstein GAN~\cite{arjovsky2017wasserstein} with gradient penalty, was employed to directly enhance distorted underwater images. Subsequently, Islam et al.~\cite{islam2020fast} proposed a network called FUnIE-GAN. In order to supervise the adversarial training process, they designed several loss functions based on global content, color, local texture, and style information. In~\cite{li2018emerging}, a weakly supervised network based on CycleGAN was applied to the UIE task. It maps color and style of underwater images to those of above-water natural images, thereby correcting the color cast in underwater images.

\subsection{Other deep learning-based UIE methods}

Differing from CNN, which adopts local receptive kernels to extract features from images, Vision Transformer (ViT)~\cite{dosovitskiy2020image} treats an image as a sequence composed of multiple patches and uses self-attention computation to capture global information of the image. Inspired by this, Peng et al.~\cite{peng2023ushape} introduced Transformer into UIE for the first time. They proposed a U-shape Transformer network consisting of a channel-wise multi-scale feature fusion transformer module and a spatial-wise global feature modeling transformer. In addition, some UIE methods have reexamined physical models and combined deep neural networks with them. SyreaNet~\cite{wen2023syreanet} integrated both synthetic and real-world data under the guidance of a revised underwater image formation model to fulfil the enhancement aim. Furthermore, apart from using physical models to guide the UIE models, there are also methods aiming to leverage high-level information to guide learning process of the network. In~\cite{qi2022sguie}, they managed to use information from semantic segmentation to force network to learn region-wise enhancement features from underwater images and their paired high-quality reference images.

\section{Methodology}\label{sec.3}
\subsection{RAUNE-Net}
\begin{figure}
    \centering
    \includegraphics[width=0.85\textwidth]{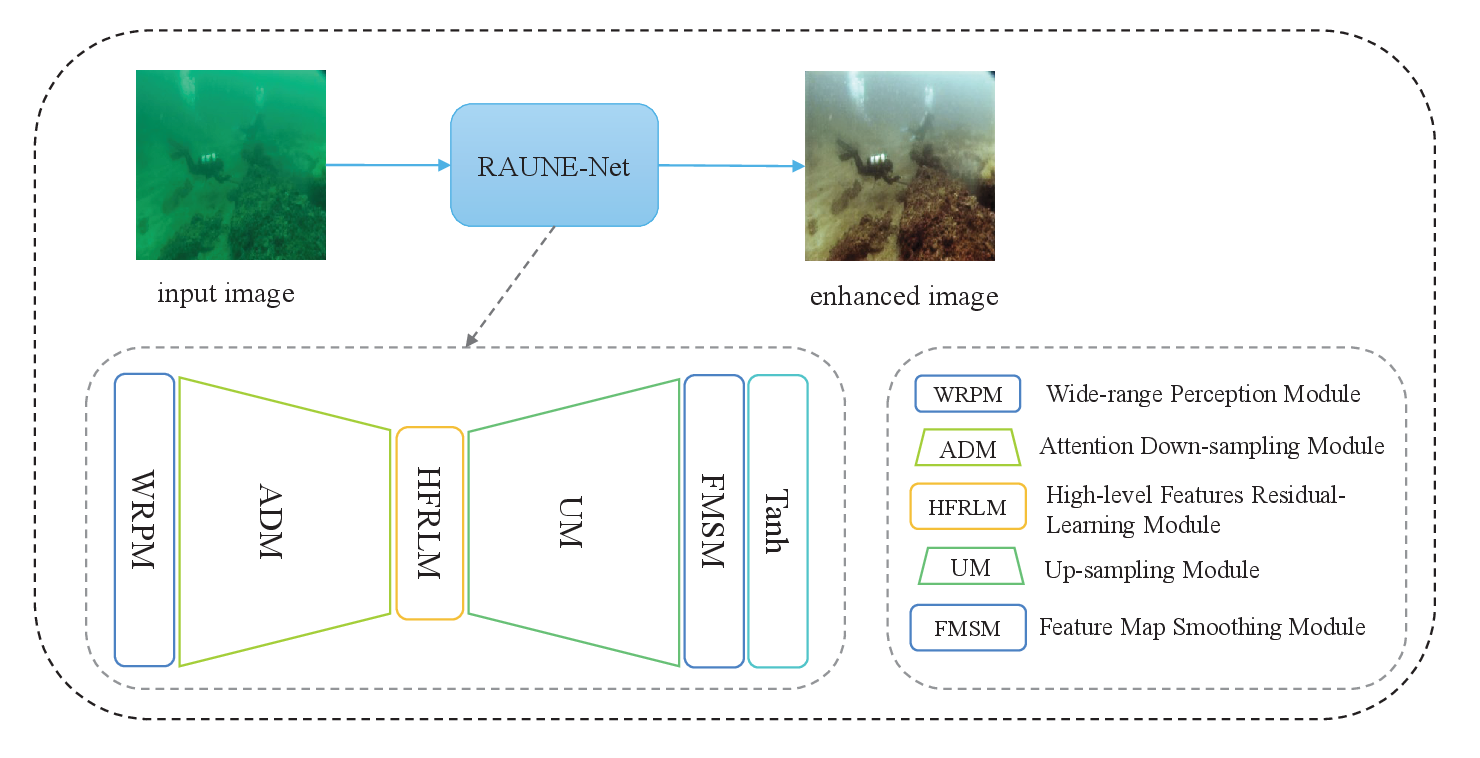}
    \caption{The data-flow and architecture of RAUNE-Net}
    \label{raune-net}
\end{figure}

We have observed significant improvements in objective metrics for some deep neural networks that utilized residual~\cite{he2016deep} and attention~\cite{woo2018cbam} mechanisms. Consequently, we combine residual modules and attention blocks to design the RAUNE-Net. As demonstrated in Figure \hyperref[raune-net]{2}, our network follows an end-to-end manner. Given a raw underwater image, by feeding it through the network, we can obtain an enhanced result. Mathematically, this enhancement process can be written as:

\begin{equation}
    Y_e = \Phi(X_i),
\end{equation}

\noindent where $X_i$ are input underwater images, $\Phi$ is RAUNE-Net, and $Y_e$ are enhanced outputs.

\begin{figure}
    \centering
    \includegraphics[width=0.95\textwidth]{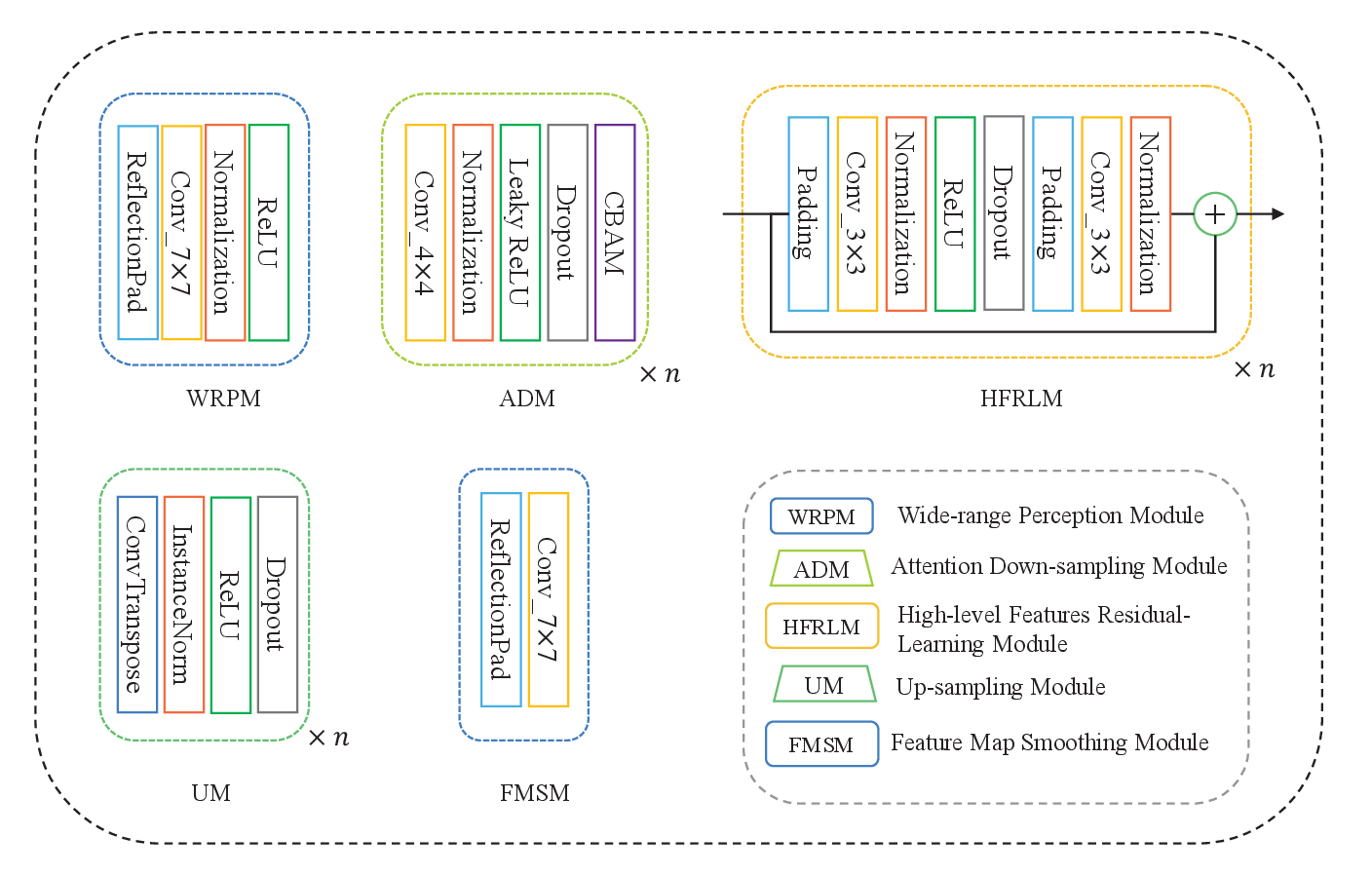}
    \caption{The detail structure of RAUNE-Net}
    \label{raune-net-details}
\end{figure}

Overall, the RAUNE-Net consists of a wide-range perception module (WRPM), an attention down-sampling module (ADM), a high-level features residual-learning module (HFRLM), an up-sampling module (UM), and a feature map smoothing module (FMSM) followed by a Tanh activation layer. The detail structure of our network is illustrated in Figure \hyperref[raune-net-details]{3}.

In detail, the WRPM contains a reflection padding layer, a convolutional layer, a normalization layer, and a non-linear activation layer. Specifically, this convolutional layer has a large kernel size of $7\times7$, which can cover a larger receptive field, enabling the network to capture more extensive contextual information, identify the overall structure and important features in the image. We set the default number of convolutional kernels to 64, use Instance Normalization~\cite{ulyanov2016instance} as the default normalization layer, and employ ReLU as the activation function.

ADM consists of multiple blocks, each block containing a $4\times4$ kernel size convolutional layer with a stride of 2 and a padding size of 1, followed by a normalization layer, a LeakyReLU~\cite{maas2013rectifier} activation layer, an optional dropout~\cite{srivastava2014dropout} layer, and an attention module. Specifically, the attention module in ADM combines channel attention calculation and spatial attention calculation in a sequential manner~\cite{woo2018cbam}. After down-sampling, HFRLM consists of several residual blocks, where each residual block contains two convolutional layers with padding layers placed before them and normalization layers assigned after them. A ReLU activation layer and an optional dropout layers are inserted between the two convolutional layers to prevent overfitting and improve the network's robustness.

Similar to down-sampling stage, the UM is composed of multiple up-sampling blocks. Each up-sampling block includes a convolutional transpose layer with a kernel size of $4\times4$, stride of 2, and padding size of 1, followed by a normalization layer, a non-linear activation layer, and an optional dropout layer. After up-sampling, we do not directly output the result. Instead, we add a module called FMSM, which consists of a reflection padding layer and a convolutional layer with a kernel size of $7\times7$. We adopt this module because we have observed that the enhanced images often exhibit varying degrees of checkerboard artifacts. By performing a convolution operation with a large kernel, we can smooth the feature maps and reduce the occurrence of jagged edges and noise. This makes the up-sampled images more natural, thereby improving the visual quality~\cite{liu2015frame,liu2020blind} of the output. Lastly, we incorporate a Tanh non-linear activation layer to map the pixel values of the image to the range of -1 to 1, enhancing its contrast and color variation range.

\subsection{Loss Function}
To train our model, we use a weighted combination of three types of loss functions as the overall loss, i.e., the pixel content loss $\mathcal{L}_{pcont}$, the structural similarity loss $\mathcal{L}_{ssim}$, and the semantic content loss $\mathcal{L}_{scont}$:

\begin{equation}
    \mathcal{L} = \lambda_{pcont} \cdot \mathcal{L}_{pcont} + \lambda_{ssim} \cdot \mathcal{L}_{ssim} + \lambda_{scont} \cdot \mathcal{L}_{scont}.
\end{equation}

\subsubsection{Pixel Content Loss}
The pixel content loss is designed to measure the discrepancy in pixel values between the output image $Y_{e}$ generated by the model and the reference image $Y_{ref}$ across the red, green, and blue color channels. Here, to reduce computational cost, we utilize the absolute distance between pixel values to represent the difference between the two images. The mathematical formula is defined as:

\begin{equation}
    \mathcal{L}_{pcont} =  \mathbbm{E}\left[{\left\| Y_{e} - Y_{ref} \right\|}_{1}\right].
\end{equation}

\subsubsection{Structural Similarity Loss}
Structural Similarity Index Measure (SSIM)~\cite{wang2004image,liu2017reduced} is a metric that quantifies the structural similarity between two images based on luminance, contrast, and structural information. The main purpose of incorporating this metric into the loss function is to encourage the network to generate enhanced images that are closer to the ground truth in these three aspects. Specifically, the value of SSIM ranges from 0 to 1, and a value closer to 1 indicates a higher similarity to the reference image. However, typically in optimization, the value of objective function is minimized. Therefore, during the training process, the expression for SSIM Loss is:

\begin{equation}
    \mathcal{L}_{ssim} = \mathbbm{E}\left[ \frac{1 - SSIM\left(Y_{e}, Y_{ref}\right)}{2} \right],
\end{equation}

\noindent where $Y_{e}$ are enhanced images, $Y_{ref}$ are reference images, and the $SSIM\left(*,*\right)$~\cite{wang2004image} represents structural similarity index map between two images. Additionally, the index map is calculated as:

\begin{equation}
    SSIM(x, y) = \frac{\left( 2\mu_{x}\mu_{y} + c_{1} \right) \left( 
2\sigma_{xy} + c_{2} \right)}{\left( \mu_{x}^{2} + \mu_{y}^{2} + c_{1} \right) \left( \sigma_{x}^{2} + \sigma_{y}^{2} + c_{2} \right)},
\end{equation}

\noindent where the $\mu_{x}$ and  $\mu_{y}$ represent the pixel sample mean of image $x$ and image $y$, the $\sigma_{y}^{2}$, $\sigma_{y}^{2}$ and $\sigma_{xy}$ stand for variances and covariance of $x$ and $y$, and $c_{1}$ and $c_{2}$ are two variables to stabilize the division with weak denominator, respectively.

\subsubsection{Semantic Content Loss}
\begin{figure}
    \centering
    \includegraphics[width=0.8\textwidth]{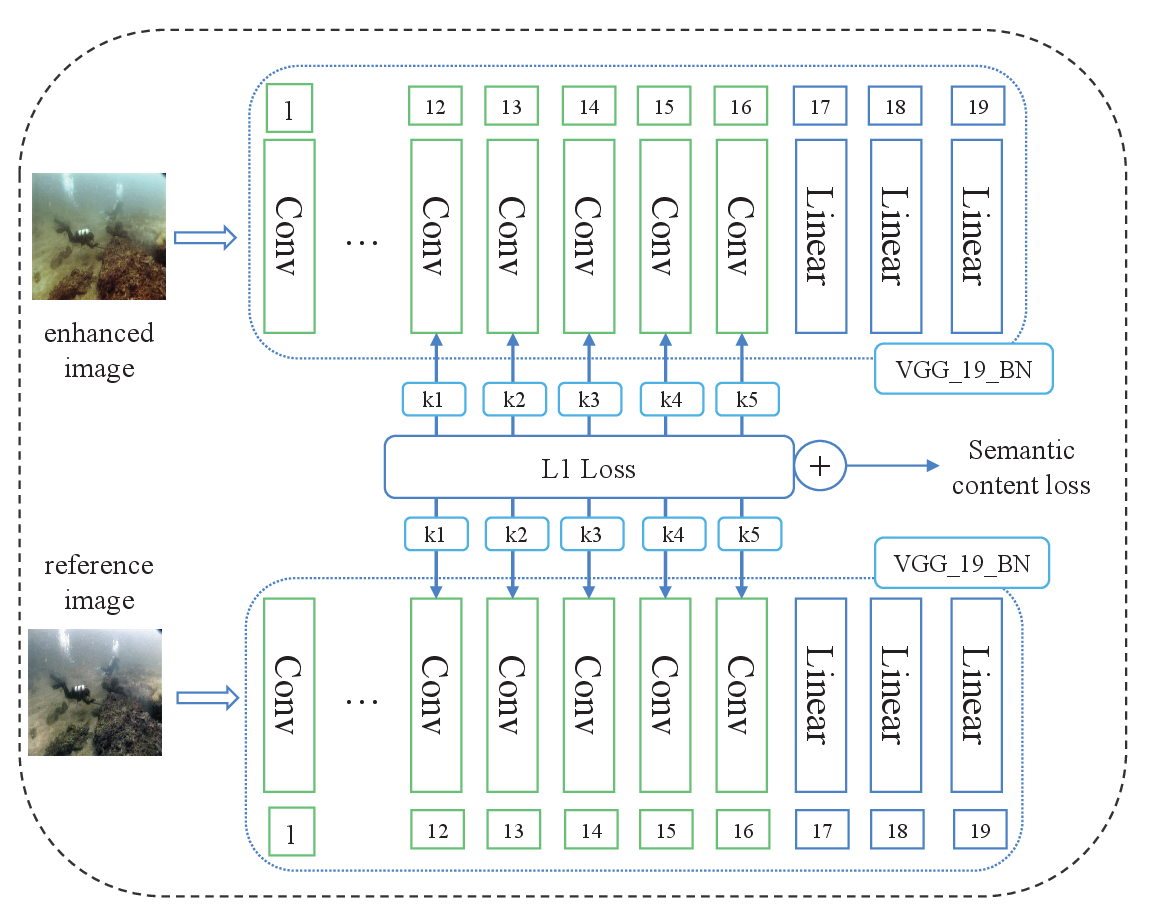}
    \caption{Diagram of the semantic content loss}
    \label{scont-loss}
\end{figure}

 In networks used for image classification, the features obtained from deeper convolutional layers to some extent represent the semantic content of the input image. Therefore, we consider extracting high-level features obtained from last several layers of a classification network to evaluate the semantic discrepancy between the enhanced image and the reference image. Specifically, we utilize the pre-trained VGG19\_BN~\cite{simonyan2014very} model trained on the ImageNet-1k\_v1~\cite{imagenet15russakovsky} dataset. As shown in Figure \hyperref[scont-loss]{4}, we extract feature maps obtained from the last five convolutional layers for both the enhanced image and the reference image. Finally, we calculate the weighted sum of the L1 distances between each pair of feature maps, which serves as the semantic content loss. Mathematically, this loss can be written as:

\begin{equation}
    \mathcal{L}_{scont} = \sum_{i=1}^{5}{k_{i} \cdot MAE(\Omega_{i}(Y_{e}), \Omega_{i}(Y_{ref}))},
\end{equation}
\noindent where $k_{i}$ is the $i$-th weight for summing, $MAE(*,*)$ is the Mean Absolute Error (i.e., L1 loss), and the $\Omega_{i}(*)$ is the last $i$-th convolutional layer in VGG19\_BN.

\section{Experiments}\label{sec.4}

\subsection{Datasets}
Firstly, we select the LSUI3879 dataset for training, which consists of 3879 real-world underwater images and their corresponding reference images randomly partitioned from the LSUI~\cite{peng2023ushape} dataset. For testing, we use four different datasets, namely LSUI400 (the remaining 400 pairs of images from LSUI), UIEB100 (100 pairs of images randomly selected from UIEB~\cite{li2020underwater}), OceanEx, and EUVP\_Test515 (515 pairs of test samples from EUVP~\cite{islam2020fast}). In particular, OceanEx is a testset collected and constructed by ourselves. We gather 40 high-quality underwater images from \href{https://oceanexplorer.noaa.gov/image-gallery/welcome.html}{NOAA Ocean Exploration}. Then, we apply CycleGAN~\cite{Zhu_2017_ICCV} to add underwater distortions and degradation styles to these images, making them the samples to be enhanced, while keeping the original high-quality images as reference images. These four datasets are used for objective evaluation of networks' performance, with the evaluation metrics being peak signal-to-noise ratio (PSNR) and structural similarity (SSIM). In addition to that, we also employ three datasets for subjective evaluation. They are U45~\cite{li2019fusion}, RUIE\_Color90 (90 images randomly selected from the UCCS subset of RUIE~\cite{liu2020real} dataset), and UPoor200. Among them, UPoor200 is a dataset we collected and curated from the internet, which consists of 200 real-world underwater images with poor visual quality. It includes distortions~\cite{liu2019unsupervised} such as blue-green color cast, low lighting, blurriness~\cite{liu2017quality}, noise, and haze.

\subsection{Implementation Details}
Before training, the input images are first resized to a resolution of $256\times256$ and then normalized. The normalization is performed by setting the mean and standard deviation to 0.5. During training, the optimizer we used is Adam with a learning rate of 0.0001. The two coefficients (i.e, betas) of this optimizer are set to 0.9 and 0.999 respectively. The weights assigned to pixel content loss, SSIM loss, and semantic content loss are 1.0, 1.0, and 1.0 respectively. During each training process, the network is trained for 100 epochs, and the trainable parameters are saved every five epochs. Every 500 iterations, the enhanced results of four sampled images from the testset are displayed. We conducted a series of experiments on the number of attention modules and residual modules in RAUNE-Net. Additionally, we compared RAUNE-Net with eight other UIE methods (i.e., UT-UIE~\cite{peng2023ushape}, SyreaNet~\cite{wen2023syreanet}, WaterNet~\cite{li2019underwater}, UGAN~\cite{fabbri2018enhancing}, FUnIE-GAN~\cite{islam2020fast}, UWCNN~\cite{li2020underwater}, SGUIE-Net~\cite{qi2022sguie}, Cycle-GAN~\cite{li2018emerging}). To ensure a fair comparison, we retrained some networks mentationed in these methods (i.e., WaterNet, UGAN, and FUnIE-GAN) using the LSUI3879 dataset.

\subsection{Objective Evaluations}
The results of objective evaluation~\cite{liu2018blind} of nine UIE methods, including RAUNE-Net and other eight UIE methods compared to it, are demonstrated in Tabel \hyperref[tab2]{1}. From the table, our RAUNE-Net achieves highest PSNR values and SSIM values, which are marked as red color, on LSUI400, EUVP\_Test515 and UIEB100 testsets. On OceanEx, RAUNE-Net achieves the highest PSNR value and the second highest SSIM value, but the SSIM value is already very close to the highest value obtained by WaterNet. It is worth noting that WaterNet achieves slightly lower performance values (except for SSIM value on OceanEx), which are indicated with blue color, than RAUNE-Net on the four test sets. The reason for the improved performance after retraining is the use of SSIM loss and the larger training dataset (i.e., LSUI), which contains more diverse underwater scenes. The UWCNN consistently achieves the lowest objective values on all four datasets. This is mainly due to its training on synthesized images using a physical model, which results in weak generalization and poor adaptability.

\begin{table}
    \centering
    \caption{Objective evaluation results of different UIE methods}
    \label{tab2}
    \begin{tabular}{ccccccccc}
        \toprule
        \multirow{2}{*}{Methods} & \multicolumn{2}{c}{LSUI400} & \multicolumn{2}{c}{EUVP\_Test515} & \multicolumn{2}{c}{UIEB100} & \multicolumn{2}{c}{OceanEx} \\
        \cmidrule(lr){2-3} \cmidrule(lr){4-5} \cmidrule(lr){6-7} \cmidrule(lr){8-9}
        & PSNR & SSIM & PSNR & SSIM & PSNR & SSIM & PSNR & SSIM \\
        \midrule
        UT-UIE~\cite{peng2023ushape} & 24.351 & 0.829 & 25.214 & 0.813 & 20.916 & 0.764 & 21.270 & 0.822 \\
        SyreaNet~\cite{wen2023syreanet} & 18.050 & 0.766 & \underline{17.721} & 0.743 & 16.501 & 0.836 & 20.243 & 0.865 \\
        WaterNet~\cite{li2019underwater} & \color{blue}26.688 & \color{blue}0.874 & \color{blue}25.285 & \color{blue}0.833 & \color{blue}22.279 & \color{blue}0.868 & \color{blue}22.132 & \color{red}0.887 \\
        UGAN~\cite{fabbri2018enhancing} & 25.117 & 0.846 & 23.636 & 0.805 & 21.368 & 0.825 & 22.436 & 0.822 \\
        FUnIE-GAN~\cite{islam2020fast} & 23.272 & 0.818 & 24.077 & 0.794 & 19.614 & 0.813 & 20.448 & 0.855 \\
        UWCNN~\cite{li2020underwater} & \underline{17.366} & \underline{0.725} & 17.725 & \underline{0.704} & \underline{14.155} & \underline{0.686} & \underline{15.960} & \underline{0.724} \\
        SGUIE-Net~\cite{qi2022sguie} & 19.910 & 0.819 & 19.187 & 0.760 & 21.178 & 0.872 & 18.677 & 0.834 \\
        Cycle-GAN~\cite{li2018emerging} & 18.320 & 0.749 & 17.963 & 0.709 & 17.714 & 0.758 & 21.007 & 0.828 \\
        RAUNE-Net & \color{red}26.812 & \color{red}0.876 & \color{red}26.331 & \color{red}0.845 & \color{red}22.751 & \color{red}0.879 & \color{red}22.728 & \color{blue}0.876 \\
        \bottomrule
    \end{tabular}
\end{table}

\subsection{Subjective Evaluations}
\begin{figure}
    \centering
    \includegraphics[width=\textwidth]{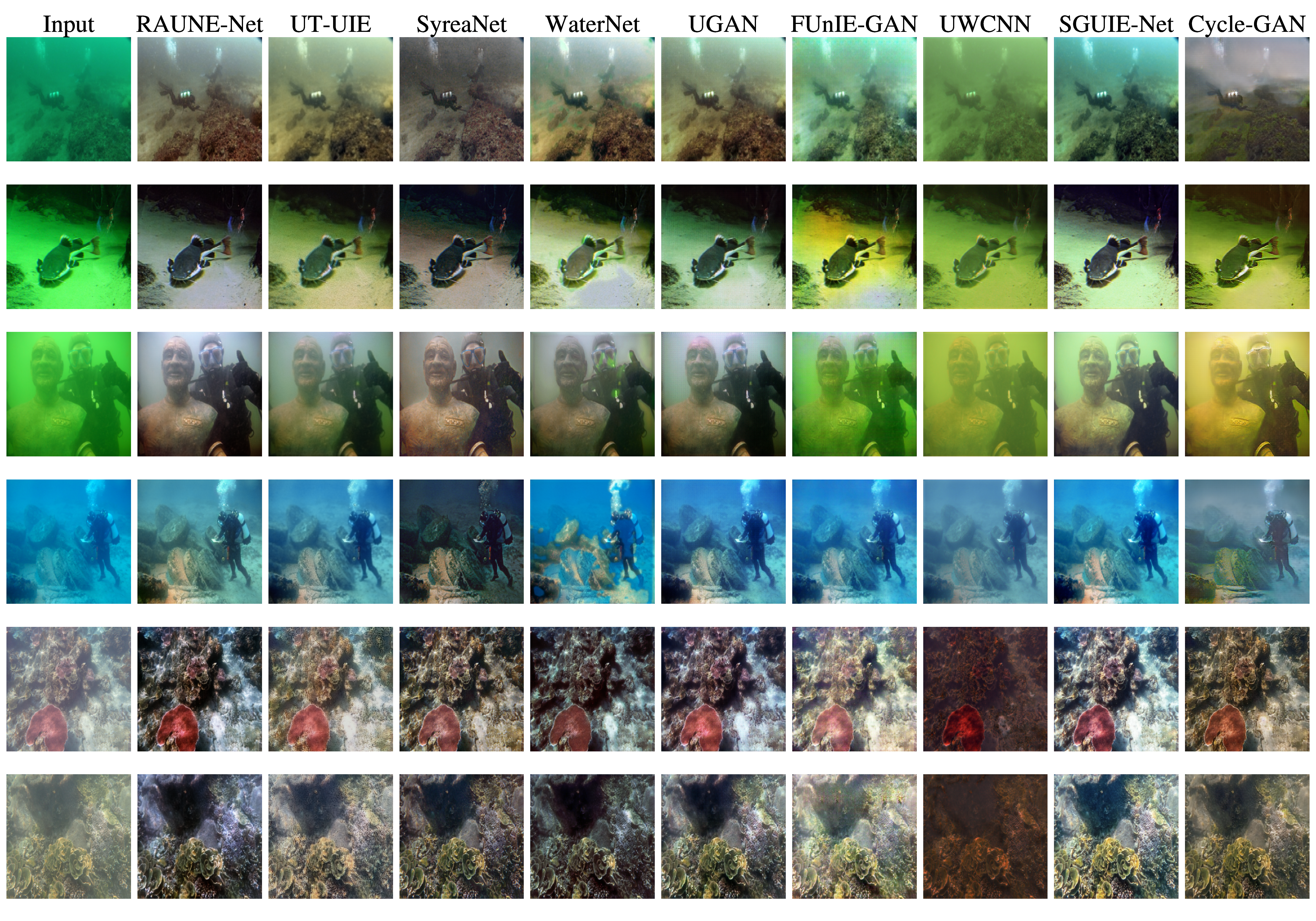}
    \caption{Subjective evaluation results of different methods on U45}
    \label{different-methods-noref-U45}
\end{figure}

As can be observed from Figure \hyperref[different-methods-noref-U45]{5}, RAUNE-Net is capable of enhancing bluish, greenish and hazy underwater images, while the other methods have exhibited improper or insufficient handling of these images. Specifically, UT-UIE and FUnIE-GAN incompletely remove the green color tone when processing the second and third images, resulting in noticeable green color residue in the results. Although SyreaNet effectively eliminates the blue-greenish color tone and foggy effect, it reduces the clarity and brightness of the source images. From the first four images, we can clearly see that WaterNet only partially removes the blue-green bias, as there are still blue or green artifacts preserved at the edges of objects in the images. UGAN, while not showing unpleasant artifacts, fails to effectively remove the blue and green color bias. For example, in the second result image, a large area of green color is retained, and in the fourth result image, the blue color style is still prominent.

\begin{figure}
    \centering
    \includegraphics[width=\textwidth]{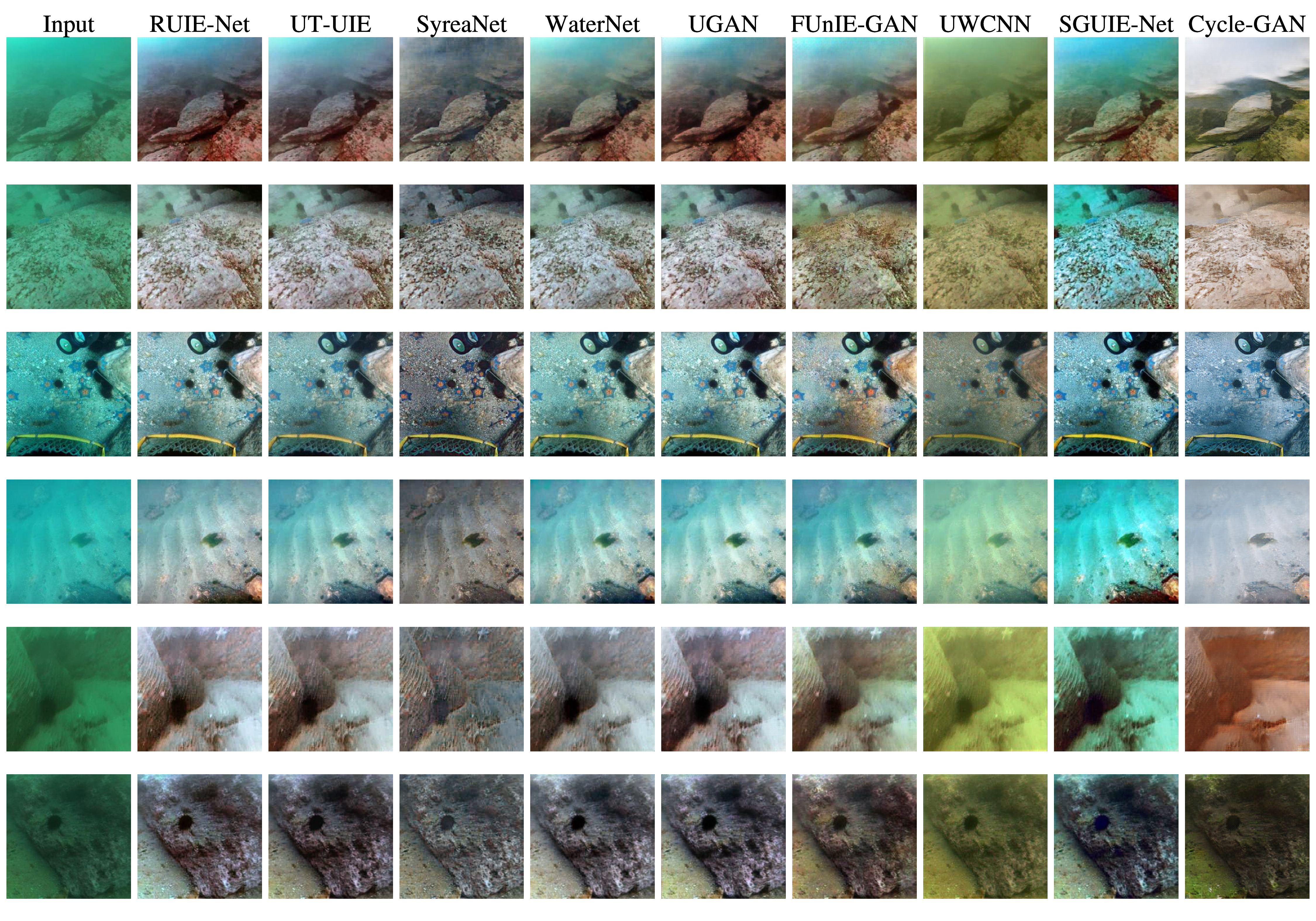}
    \caption{Subjective evaluation results of different methods on RUIE\_Color90}
    \label{different-methods-noref-RUIE-Color90}
\end{figure}

\begin{figure}
    \centering
    \includegraphics[width=\textwidth]{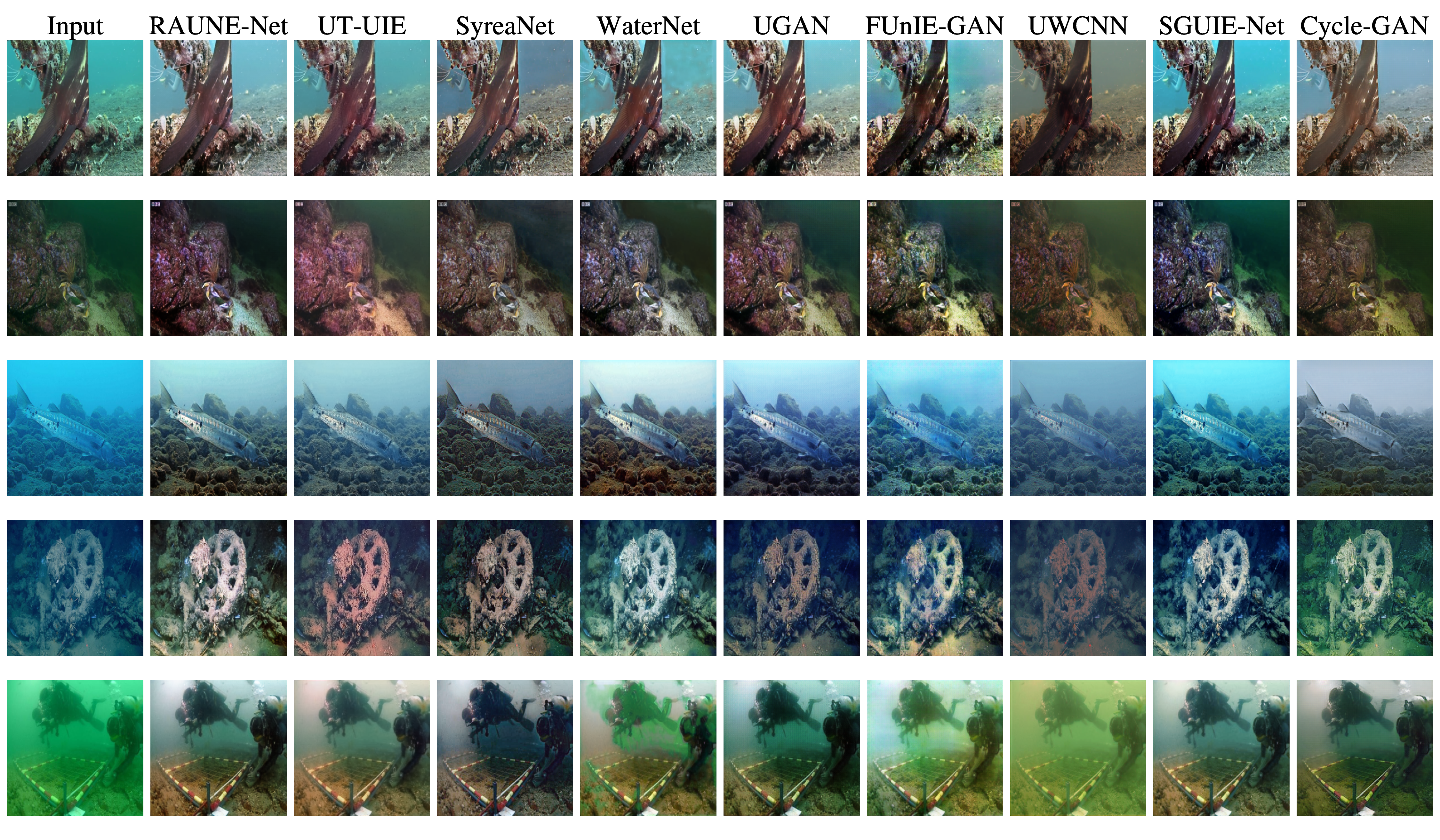}
    \caption{Subjective evaluation results of different methods on UPoor200}
    \label{different-methods-noref-UPoor200}
\end{figure}

Unfortunately, UWCNN completely fails to enhance the original image and instead makes the input image appear more turbid and with reduced visibility. This is particularly evident in the last two result images. Similarly, SGUIE-Net's ability to handle color shifts is not as impressive as that of RAUNE-Net. CycleGAN also failed to properly handle the green color shift and introduced strange textures that altered the original image content. Furthermore, as shown in Figure \hyperref[different-methods-noref-RUIE-Color90]{6} and \hyperref[different-methods-noref-UPoor200]{7}, we can observe consistent results from subjective evaluations on other two testsets. All in all, compared to other eight enhancement methods, RAUNE-Net coherently achieves remarkable visual effects.

\section{Conclusion}\label{sec.5}
In this paper, we devised a novel underwater image enhancement structure named \textit{RAUNE-Net} driven by residual learning, channel attention and spatial attention. We found that the residual learning on high-level features can help the network to learn more detail information to enhance input images and the channel attention and spatial attention blocks connected sequentially can assist the enhancing system to notice important areas to be enhanced and how to enhance them. By employing objective and subjective evaluations on seven testsets, we discovered that our network obtain an outstanding and reliable enhancing capability on various underwater attenuation types, such as bluish or greenish color shift, hazy or foggy distortion, and low-light condition. To contrive a fair and generalized assessment for different UIE methods' performances, we also constructed two datasets (a set with reference images called OceanEX and the other set with no ground-truth images named UPoor200), which contains diverse kinds of real-world underwater images. Furthermore, we found using PSNR and SSIM as indicators of the objective performance of enhanced images is still inaccurate, a non-reference quality assessment of these enhanced images is a focal point of our future research.

\bibliographystyle{splncs04}
\bibliography{my_references}

\end{document}